\documentclass{INTERSPEECH2023}
\usepackage{subfig}
\usepackage{booktabs}

\interspeechcameraready

\usepackage{todonotes}
\title{On the N-gram Approximation of Pre-trained Language Models}
\name{Aravind Krishnan$^{1,2}$, Jesujoba O. Alabi$^{1,3}$, Dietrich Klakow$^1$}
\address{
  $^1$Spoken Language Systems Group, Saarland University, Germany \\
  $^2$German Research Center for Artificial Intelligence (DFKI)\\
  $^3$Saarland Informatics Campus, Germany}
\email{\{akrishnan|jalabi|dietrich\}@lsv.uni-saarland.de}

\begin{document}

\maketitle
 
\begin{abstract}
Large pre-trained language models (PLMs) have shown remarkable performance across various natural language understanding (NLU) tasks, particularly in low-resource settings. Nevertheless, their potential in Automatic Speech Recognition (ASR) remains largely unexplored. This study investigates the potential usage of PLMs for language modelling in ASR. We compare the application of large-scale text sampling and probability conversion for approximating GPT-2 into an n-gram model. Furthermore, we introduce a vocabulary-restricted decoding method for random sampling, and evaluate the effects of domain difficulty and data size on the usability of generated text. Our findings across eight domain-specific corpora support the use of sampling-based approximation and show that interpolating with a large sampled corpus improves test perplexity over a baseline trigram by 15\%. Our vocabulary-restricted decoding method pushes this improvement further by 5\% in domain-specific settings.
\end{abstract}

\section{Introduction}


The advent of Pre-trained Language Models (PLMs) such as BERT~\cite{devlin-etal-2019-bert} and GPT-2~\cite{radford_language_2019} has led to significant strides in language processing tasks, particularly in low-resource scenarios. These models outperform conventional neural language models such as RNNLMs and transformer-based language models (LMs) through the pre-train/fine-tune paradigm. Nevertheless, the potential of these models in automatic speech recognition (ASR) has been insufficiently investigated. Within existing literature, several studies  examine their use in lattice rescoring~\cite{dingliwal2021domain, udagawa2022effect,zheng2021adapting}, but their utilization in first-pass decoding remains under-explored. This is justified to some extent, since using PLMs directly in first-pass decoding would suffer from the same limitation as using RNNLMs, which is computational inefficiency~\cite{varia-anoop,connect-schwenk}.\\
Despite the computational bottleneck, the use of RNNLMs in ASR has been studied extensively. In the context of first-pass decoding, two methods using RNNLMs have been explored, namely offline and online methods~\cite{singh2017approximated,tarjan2020effectiveness}. The offline method involves approximating an RNNLM to generate an n-gram LM that is used for first-pass decoding and the online approach employs a cache to store the RNNLM states and prunes them during decoding~\cite{li2018recurrent}. Several offline approaches to approximate RNNLMs to n-grams have been explored previously~\cite{varia-anoop, singh2017approximated,tarjan2020effectiveness,adel2014comparing}, but to the best of our knowledge, only the study by~\cite{transcov-wang} looks at a similar approximation for PLMs. Their approach involves pre-training a GPT-like model on a very large general-domain corpus, fine-tuning it on a smaller domain-specific corpus, and then using it to generate a large corpus for approximation into an n-gram LM. They use pre-training corpora that have hundreds of thousands of sentences but approximation has not been explored in a low-resource scenario, where trends can vary significantly in comparison. Moreover, these approximation methods have not been studied for GPT-2, which is much larger and more powerful than GPT and has shown impressive few- and zero-shot capabilities~\cite{radford_language_2019} in comparison. 
Hence, this paper explores the use of GPT-2 in approximating and augmenting n-gram LMs under low-resource scenarios. Specifically, we compare text sampling-based conversion and probability-based conversion as a means of approximating GPT-2 in domain-specific settings. Our research contributions are: 
\begin{enumerate}
    \item Comparing the efficiency of text sampling and probability conversion in approximating GPT-2 into an n-gram model.
    \item Introducing a vocabulary-restricted decoding method that improves text generation in domain-specific scenarios.
    \item Evaluating the competitiveness of sampling-based approximation across domain difficulty and data scarcity.
\end{enumerate}
\vspace{-0.2cm}
\section {Data}
\textbf{Taskmaster-2} is a spoken dialogue corpus \cite{byrne2019taskmaster} of 17,289 conversations between users and call-center operators. The data spans human conversations in six domains: restaurant-search, food-ordering, sports, music, movies, and flights. For our experiments, we combine  all utterances in a call into a single sample. A train:dev:test split of 70:20:10 is used. \textbf{HUB4 1996}~\cite{graff19971996} is a broadcast news speech corpus that contains 104 hours of transcribed speech from television networks. To simulate a low-resource scenario, we use transcripts from the speech data (LDC97T22) for training and the standard development and test datasets (LDC97S66). \textbf{ATCO2} is an air traffic domain corpus~\cite{zuluaga2022ATCO2} that contains air traffic communication between controllers and pilots. We use the 4-hour test set for training and the publicly released 1-hour test set for testing. We choose these datasets to experiment with different domains, task difficulty, and resource availability. 
The sentence and word level distribution across all datasets is shown in Table \ref{table:datasets}.
\begin{table}[!h]
\centering
\begin{tabular}{lccc}
    \toprule
    Dataset& \multicolumn{3}{c}{\#Sentences \textbf{/} \#Words} \\
    \cmidrule(lr){2-4} 
    & Train & Dev & Test \\
    \midrule
    ATCO2 & 2337 \textbf{/} 27K  & 537 \textbf{/} 6K & 826 \textbf{/} 10k \\
    HUB4 & 46573 \textbf{/} 735K & 4178 \textbf{/} 65K & 346 \textbf{/} 19K\\
    Movies & 2139 \textbf{/} 344K & 612 \textbf{/} 97K & 305 \textbf{/} 49K \\
    Restaurant-Search & 2293 \textbf{/} 351k & 656 \textbf{/} 98k & 327 \textbf{/} 49k\\
    Music & 1122 \textbf{/} 142K & 321 \textbf{/} 40K & 160 \textbf{/} 19K\\
    Flights & 1736 \textbf{/} 332K & 497 \textbf{/} 96K & 248 \textbf{/} 47K\\
    Sports & 2436 \textbf{/} 279K & 697 \textbf{/} 81K  & 348 \textbf{/} 39K\\
    Food-Ordering & 735 \textbf{/} 80K  & 210 \textbf{/} 22K & 105 \textbf{/} 11K\\
    \bottomrule
\end{tabular}
\caption{Dataset distribution and splits.}
\label{table:datasets}
\vspace{-0.6cm}
\end{table}

\label{approximating_gpt}

\section {Approximating GPT-2}
Existing literature supports the use of three offline methods for approximating RNNLMs into n-gram LMs: sampling-based approximation~\cite{varia-anoop}, probability-based approximation~\cite{adel2014comparing} and iterative approximation~\cite{conv-arısoy}. We concentrate on sampling and probability-based approximation methods for GPT-2 and briefly describe these methods in  Sections \ref{section:sampling} and \ref{section:probability_conversion}. We also present a vocabulary-restricted decoding scheme that controls the domain drift during text generation in Section \ref{section:vrs}.

\subsection{Sampling-Based Approximation}
\label{section:sampling}
Given a background corpus $B$, sampling-based approximation (SBA) involves generating additional text samples from the same distribution as $B$ using a language model such as an RNNLM or Transformer-based LM. The generated text is then used to train a traditional n-gram LM, which approximates the probability distribution of the neural model itself~\cite{varia-anoop}. The performance of this approach is often improved by interpolating the approximated n-gram LM with the baseline n-gram LM~\cite{singh2017approximated,adel2014comparing} trained on $B$. This approach has been studied with RNNLMs~\cite{varia-anoop,singh2017approximated,adel2014comparing} and Transformer-based LMs~\cite{transcov-wang} but has not been explored with massively pre-trained PLMs. 
In this work, we study how this approximation performs with GPT-2 across datasets, data sizes, and domains. 
\vspace{-0.1cm}
\subsection{Probability-Based Approximation}
\label{section:probability_conversion}
We also explore the use of probability-based approximation (PBA) for GPT-2, which was initially proposed by ~\cite{adel2014comparing} as a way to approximate RNNLMs into n-gram LMs. Instead of using count-based probabilities, this method involves extracting and assigning n-gram probabilities directly from trained RNNLMs. Since GPT-2 uses subword (BPE) tokens, we obtain word-level probabilities by multiplying the conditional probabilities of each BPE token in the target word. For a word $W$ composed of BPEs $\{w_1,w_2\dots w_n\}$, the word level probability is computed as:
\begin{equation}
    p(W|H) = \prod_{i=1}^n p_{gpt2} (w_i|w_{i-1}\dots w_1H)
\vspace{-0.1cm}
\end{equation}
Where $H$ is the sentence history and $p_{gpt2}$ is the conditional probability assigned to a BPE token by the fine-tuned GPT-2. The conditional word probabilities are reassigned to n-gram probabilities and averaged according to the original setup. Backing off is done from history sums taken from a trigram LM developed from the training corpus. The authors in~\cite{adel2014comparing} observe that the resulting model produces the best perplexities when unigram probabilities are borrowed from the baseline trigram LM. 


\vspace{-0.1cm}
\subsection{Vocabulary-Restricted Decoding}
\label{section:vrs}
In addition, we propose a vocabulary-constrained extension to SBA in the PLM setup. In this approach, we restrict the pool of BPE tokens that can be generated to the BPE vocabulary of the training set. We start by creating a BPE vocabulary over all words in the training data. Each word is tokenized
\footnote{GPT-2 tokenizes words with and without preceding blanks (``$apple$'' and ``$␣apple$'') differently. We add both versions into the vocabulary} 
and all composite BPEs are added to the BPE vocabulary. During random sampling,  softmax operations are performed only over the BPEs in our vocabulary. The sampling algorithm is thus restricted to picking from the BPEs present in this vocabulary. We call this Vocabulary-Restricted Decoding. \\
We distinguish our decoding scheme from lexically constrained decoding~\cite{hokamp2017lexically,hu2019improved}, where restrictions are placed on \textit{including} specific tokens during generation. Our approach uses the training set vocabulary exclusively during decoding, thus \textit{excluding} non-relevant tokens. We expect that in domain-specific scenarios, constraining the generation vocabulary will suppress tokens and force GPT-2 to model the words in the training set better. 
\label{section:preprints}

\begin{table*}[!ht]
\centering
\centerline{
\begin{tabular}{l@{\hspace{2pt}}c@{\hspace{8pt}}c@{\hspace{9pt}}c@{\hspace{8pt}}c@{\hspace{8pt}}c@{\hspace{8pt}}c@{\hspace{8pt}}c@{\hspace{8pt}}c}
    \toprule
    Model& ATCO2 & HUB4 &  \multicolumn{6}{c}{Taskmaster}  \\
    \cmidrule(lr){4-9} 
    & & &  Movies & Restaurant & Music & Flights & Sports & Food\\ 
    \midrule 
    \\
    KN3                           & \textbf{31.95} / 428 & \textbf{275.19} / 517 & \textbf{22.31} / 
    1210 & 23.01 / 1404 & 34.41 / 1269 & 19.50 / 556 & 13.24 / 639 & 12.62 / 296 \\
    RS-KN3                       & 44.00 / 446 & 310.19 / 469  & 25.67 / 1004 & 25.09 / 1241  & 31.11 / 639 & 17.59 / 397 & 12.77 / 605 & 9.96 / 119 \\
     VR-KN3         & 35.53 / 445 & 342.72 / 452 & 23.54 / 980 & \textbf{23.34} / 1066 & \textbf{28.41} / 969& \textbf{16.47} / 404& \textbf{12.35} / 593& \textbf{9.91} / 260\\

    \midrule
    \\
    KN3 + RS-KN3                   & 29.27 & \textbf{207.16}  & 19.58 & 20.26 & 26.03 & 15.96 & 11.45 & \textbf{9.33} \\
    KN3 + VR-KN3     & \textbf{28.60} & 238.21  & \textbf{19.12} & \textbf{19.71} & \textbf{25.47} & \textbf{15.70} & \textbf{11.38} & 9.53 \\
    RS-KN3 + VR-KN3 & 35.21 & 241.74 & 25.92  & 22.96 & 26.64 & 16.34 & 12.19 & 9.49 \\
    \midrule
    Total Interpolation                           & 28.54 / \textcolor{blue}{415} 
    & 207.32 / \textcolor{blue}{337} 
    & 18.91 / \textcolor{blue}{795}  
    & 19.54 / \textcolor{blue}{955} 
    & 24.38 / \textcolor{blue}{440} 
    & 15.53 / \textcolor{blue}{249} 
    & 11.29 / \textcolor{blue}{583} 
    & 9.10 / \textcolor{blue}{98 }\\
\bottomrule
\end{tabular}}
 \caption{Perplexities and OOV rates for the n-gram models constructed from sampled corpora. KN3 denotes the trigram LM baseline. Metrics are of the format: $Perplexity/\#OOVs$. All perplexities are computed using the vocabulary obtained from combining all three corpora, whose OOV rate is indicated in blue. }
\label{table:main}
\vspace{-0.4cm}
\end{table*}

\begin{table}[!t]
\centering
\begin{tabular}{lccc}
    \toprule
    Dataset& GPT-2 & RS-KN3 & PBA \\
    \midrule 
    ATCO2 & 25.04  & 47.04 & 80.04 \\
    HUB4 & 189.91 & 369.90 & 632.92\\
    Movies & 13.53 & 27.92 & 49.89 \\
    Restaurant-Search & 13.26 & 28.39 & 46.60\\
    Music & 11.55 & 35.23 & 57.83\\
    Flights & 7.23 & 17.83 & 18.10\\
    Sports & 12.77 & 13.46  & 30.74\\
    Food-Ordering & 9.96  & 10.37 & 23.31\\
\bottomrule
\end{tabular}
\caption{Sampling-based approximation of GPT-2 into an n-gram model consistently outperforms probability-based approximation}
\label{table:comparing_approximations}
\vspace{-0.7cm}
\end{table}


\section{Experiments}
\subsection{Experimental Setup}
We use the 124M parameter version of GPT-2 from  $huggingface$~\cite{wolf-etal-2020-transformers} for all experiments. The model is fine-tuned with a learning rate of 5e-5 and a weight decay of 0.01. The learning rate is linearly warmed up using $min(\#train~samples, 100)$ steps. Training is terminated using early stopping on the development set with patience of $5$. Text generation uses top-$p$~\cite{holtzman2019curious} sampling with $p$$=$$0.95$ and a temperature of $1.0$. Unless specified otherwise, we generate 100 times the training data for all SBA experiments. The n-gram models are trigram models with Kneser-Ney smoothing~\cite{chen1999empirical}, and are built and evaluated with the SRILM~\cite{stolcke2002srilm} Toolkit. \textbf{KN3} denotes the baseline model trained on just the training corpora; \textbf{RS-KN3} is trained on the random-sampled corpora and \textbf{VR-KN3} on corpora sampled using vocabulary-restricted decoding. Note that RS-KN3 and VR-KN3 are both SBA approaches with different decoding schema. All experiments use a vocabulary that combines all words in the training data and the 100x corpora generated for RS-KN3 and VR-KN3\footnote{Except in Section \ref{results:comparing_methods} where the test vocabulary is additionally introduced}. This results in an OOV rate of less than 2\% for each dataset.

\vspace{-0.1cm}
\subsection{Comparing Approximation methods}
\label{results:comparing_methods}
We compare the test set perplexities of the approximation approaches discussed with the ``true'' perplexity of a fine-tuned GPT-2 and present the results in Table \ref{table:comparing_approximations}. Word level perplexities are obtained from GPT-2  using the computation described in \cite{Mie2016Can} and elaborated in \cite{cotterell-etal-2018-languages}. To simplify, we add the test vocabulary to the n-gram models, thus having zero OOV test rate.  
Except for \textit{sports} and \textit{food-ordering}, both approximation methods incur at-least 2x degradation when compared to the true perplexity but PBA incurs significantly higher deterioration in comparison. This is consistent with previously reported results using PBA \cite{singh2017approximated,adel2014comparing}, but more drastic with GPT-2. This could be explained by PLMs' tendency to overestimate n-gram probabilities in comparison to RNNLMs. A closer look confirms that GPT-2 overestimates lower-order n-gram probabilities during PBA because of probability sharing between higher and lower-order n-grams. It is to be noted that the vocabulary set for GPT-2 is BPE based and is more fine-grained than the word-level model. We can therefore only assume that our computation for the non-approximated GPT-2 perplexity underestimates the true metric. Nevertheless, as the underestimated metric outperforms the word-based model, the granularity of the vocabulary does not interfere with our finding. We conclude that PBA is a lossy conversion scheme for GPT-2 approximation and focus on the more performant SBA method for the rest of this paper.

\subsection{Vocabulary Restriction and Interpolation}
Table \ref{table:main} compares the n-gram models developed from the training corpora and the corpora generated from random sampling and vocabulary-restricted decoding. Except for ATCO2, both sampling methods produce an OOV reduction on the test set. Combining vocabularies from the original corpus and the sampled corpora reduces the OOV count significantly across all datasets. A closer look at the OOVs contributed shows that vocabulary-restricted decoding caters to affixing and derivational inflections while random sampling has additional coverage over novel nouns and verbs.\\
In relative terms, we see that RS-KN3 under-performs the  baseline considerably (+40\% ppl) with ATCO2, where the unconventional grammatical structure adversely affects sampling. On the other hand, VR-KN3 performs strongly here with 20\% improvement over KN3. This suggests that controlling the drift of random sampling is beneficial in tightly constrained domains. In general, we see that vocabulary restriction improves over random sampling when the domain (and consequently the vocabulary) is restricted. However, performance suffers with HUB4, where the domain is wide-ranging and the vocabulary cannot be restricted as dramatically. Vocabulary restriction also outperforms random sampling for all datasets in Taskmaster, where a satisfactory equilibrium exists between grammatical complexity and domain restriction.\\
In all cases, an improvement over the baseline is observed as soon as either of the sampling methods is interpolated with the baseline KN3 (interpolation weights are tuned with the respective development sets). Here again, domain specificity plays its part, and interpolation with VR-KN3 outperforms RS-KN3 in the case of ATCO2 while the latter performs strongly for HUB4. Interpolation with either sampled model shows promising improvement with Taskmaster. We observe the best performance across all datasets when all three models are interpolated together. On average, interpolating the baseline with RS-KN3 gives a 15\% reduction in test perplexity and additional interpolation with VR-KN3 decreases perplexity further by 5\%. We conclude that interpolation with a large-scale sampled corpus is almost always beneficial for a trigram LM and is a simple and effective method to exploit PLMs for first-pass decoding.\vspace{-0.15cm}
\section{Discussion}
 \begin{figure}[!t]
  \centering
  \includegraphics[width=\linewidth]{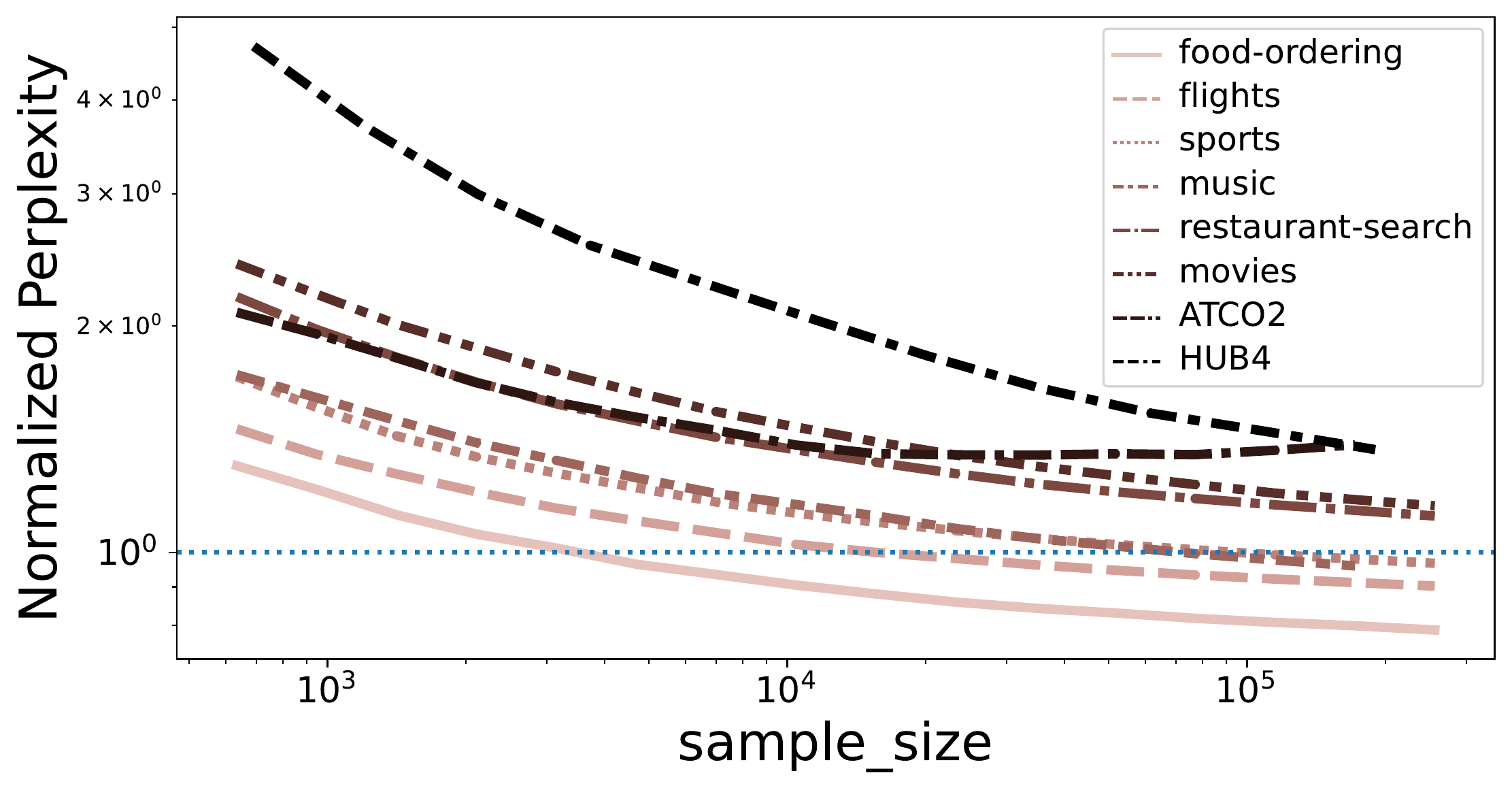}
  \caption{Benefits of sampling depends on domain difficulty. Perplexities are normalized across the respective KN3, which is shown as the horizontal line at 1.0. Darker colour denotes higher perplexity on the dev-set.\vspace{-0.3cm}}
  \label{fig:task diffculty}
  \vspace{-0.2cm}
\end{figure}
\vspace{-0.1cm}
We devote the discussion section to answering the following question - ``Can the SBA LMs outperform the baseline KN3 without interpolation?''. We isolate and study two variables that influence augmentation: the domain of the dataset and the number of training samples available. 

\subsection{Domain difficulty}

\begin{figure*}[!ht]
\vspace{-0.4cm}
    \centering
    \subfloat[Taskmaster]{
        \includegraphics[width=.325\textwidth]{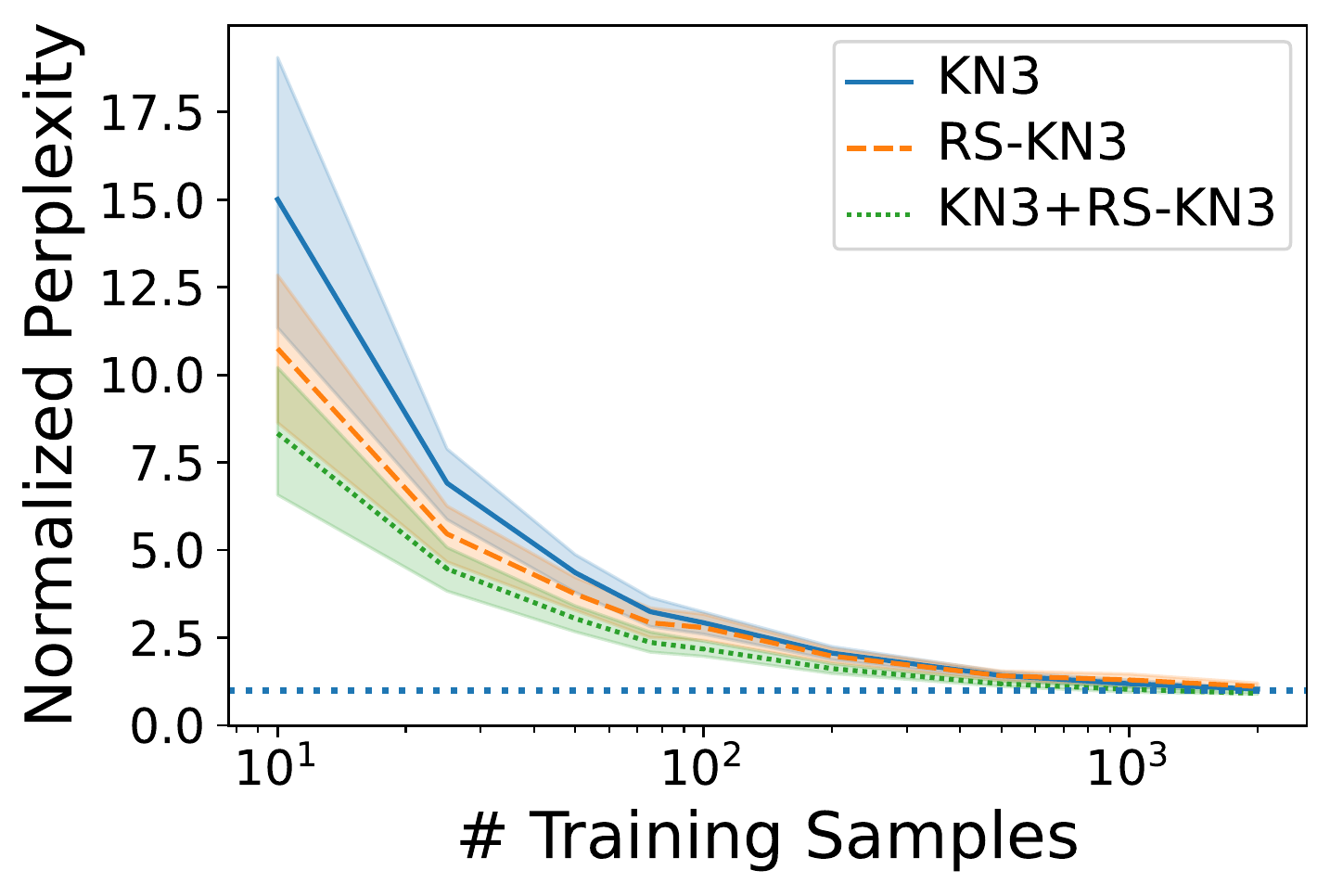}}
    \hfill
    \subfloat[ATCO2]{
        \includegraphics[width=.32\textwidth]{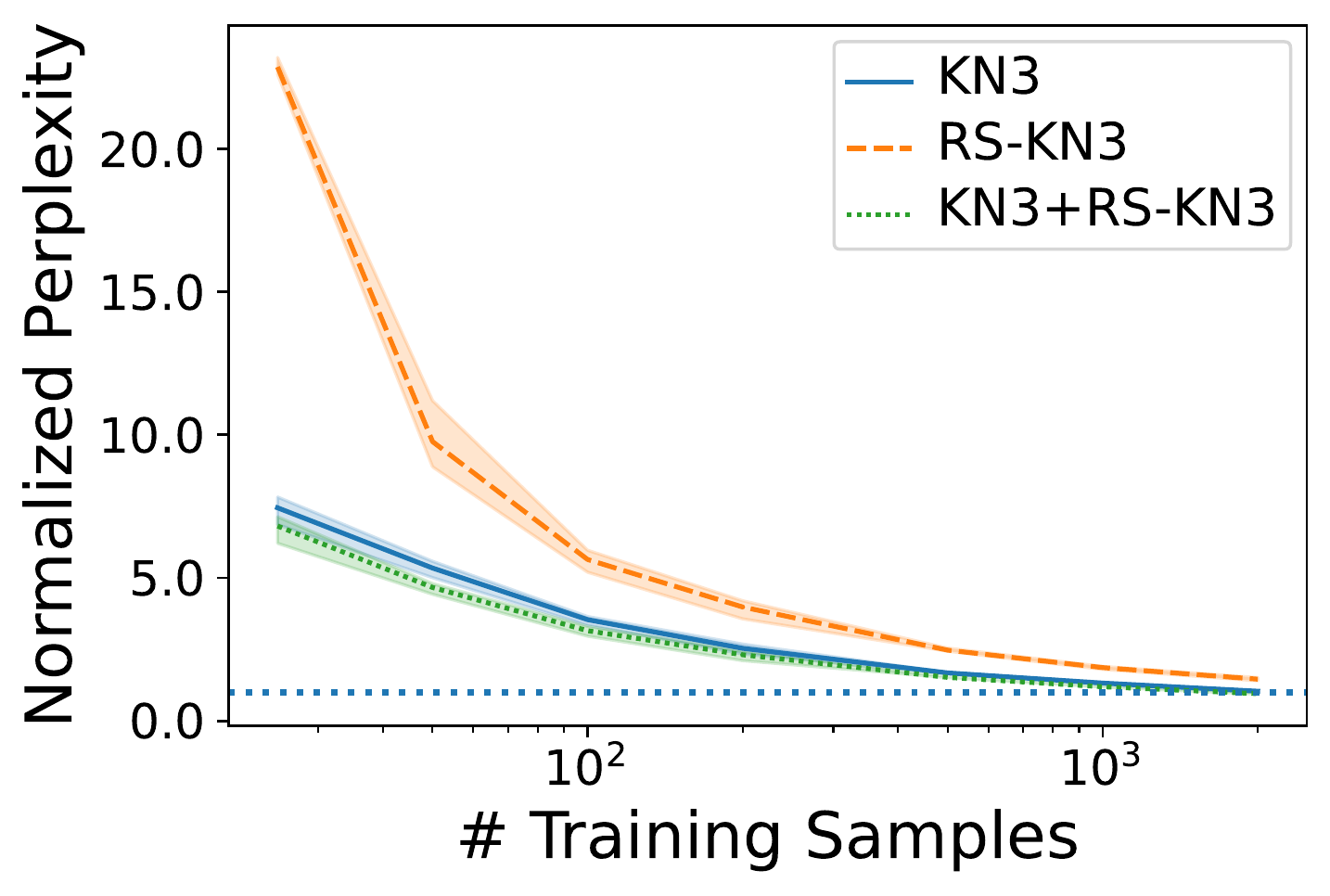}}
    \hfill
    \subfloat[HUB4]{
        \includegraphics[width=.32\textwidth]{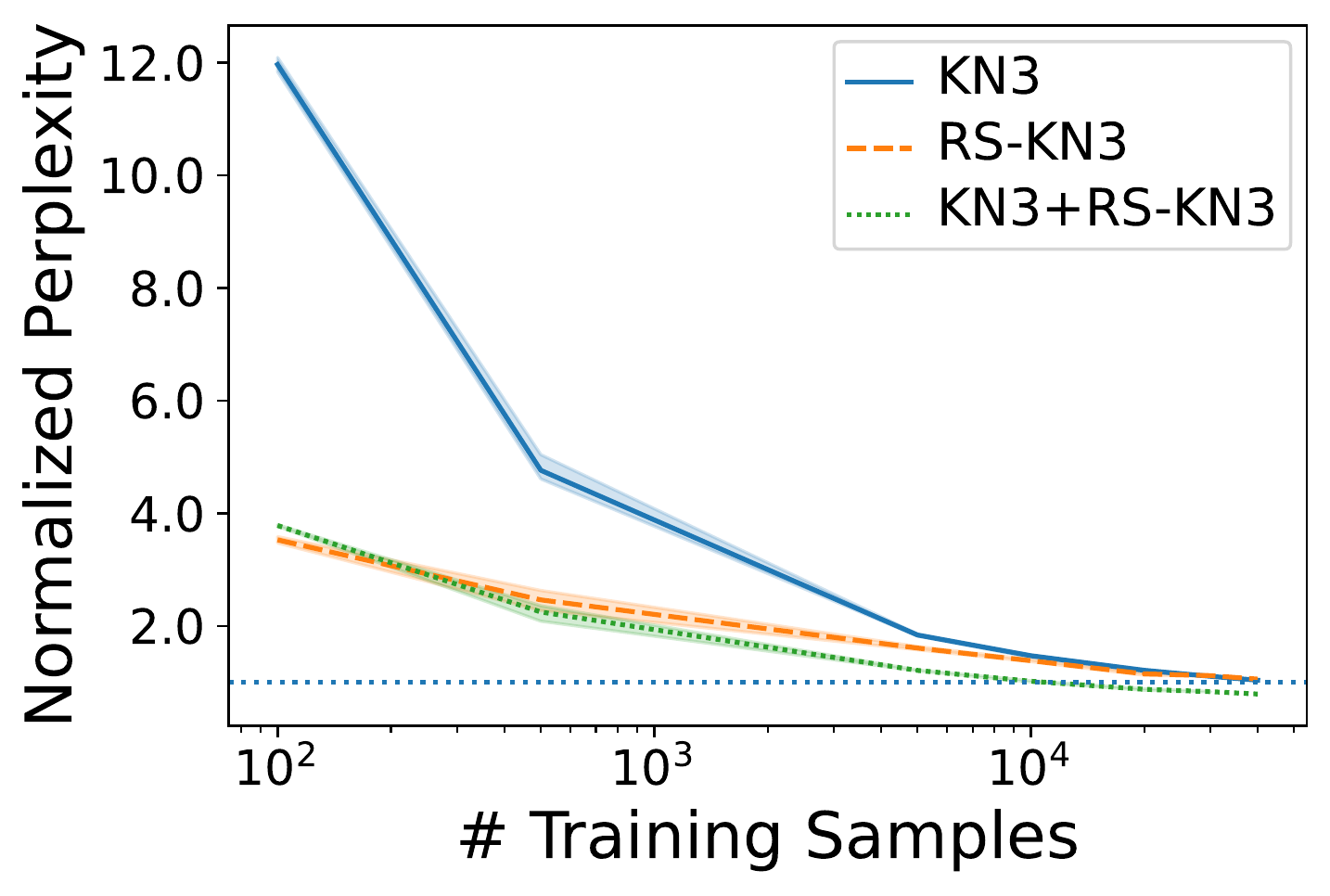}}
    
    \caption{Performance of language models across sub-sampled training corpora. Perplexities are normalized with the full-scale KN3 perplexity (which is shown as the blue dotted line at 1.0). The Taskmaster datasets are averaged to generate one plot.}
    \label{figure:train-set}
    \vspace{-0.2cm}
\end{figure*}
Comparing language modeling difficulty across two domains is challenging in a statistical setup owing to differences in vocabulary and the lack of standardized background corpora that help isolate perplexity comparisons to domains. Sub-word modeling and pre-training mitigate these concerns and PLMs prove to be ideal sandboxes to compare difficulties in modeling different domains.  Assuming that pre-training acquaints PLMs with everyday words, the adaptability of a domain over fine-tuning can then approximate the difficulty in modeling it. We quantify it by computing the dev set perplexities of fine-tuned\footnote{Perplexity from a pre-trained model could be misleading since the domain might look hard initially but be easily learnable.} GPT-2 models and compare them across domains to rank their difficulty. 
Figure \ref{fig:task diffculty} shows the trend in test perplexity with an increase in the number of sampled sentences for RS-KN3.  The figure shows that HUB4 and ATCO2 are both difficult for GPT-2 to model, but HUB4's steeper slope suggests a preference for additional sampling when domains are spread thin. In  most cases, we see that an increase in the number of samples generated correlates with a decrease in test perplexity. Larger sampling has a slower effect on perplexity after a certain point and produces long tails afterward. For datasets that have a lower dev set perplexity, SBA is seen to beat the baseline KN3 early on. An increase in domain difficulty stretches the cross-over point and we see that the number of sentences required to outmatch the baseline is proportional to the difficulty of the domain. We observe similar trends for VR-KN3 models as well, with a slight downward shift along the y-axis for all curves (except for HUB4, where the shift is slightly upwards). This analysis provides us with two conclusions: 1) Generating additional data correlates with reductions in test perplexity,
 but 2) The amount of generated data required to match the baseline trigram model has an inverse relation with domain difficulty.
\vspace{-0.1cm}
\subsection{Few-Shot Approximation}
\vspace{-0.01cm}
This section compares the responses of SBA and the baseline KN3 under low-resource settings. We simulate few-shot scenarios by sub-sampling the train and development corpora while keeping the test corpus unchanged. For each setting, GPT-2 is fine-tuned on the sub-sampled training set, using the sub-sampled development set for early stopping. The fine-tuned model then generates 100x the size of the sub-sampled train set. The resulting perplexities are normalized with the full-scale KN3 value for each dataset (refer to Table \ref{table:main}). Each experiment is run across three random seeds and the results are plotted in Figure \ref{figure:train-set}.\\
The results for ATCO2 reflect the domain difficulty yet again, with GPT-2 struggling to understand the domain in few-shot settings and the trigram model performing significantly better. GPT-2 does not outperform the trigram corpus in any data scenario, and the interpolated model hugs the KN3 curve. For Taskmaster and HUB4 on the other hand, the generated corpus fares significantly better in few-shot scenarios. The perplexity of the generated corpus is several times lower than the sub-sampled KN3 model when the number of samples available is less than 100. This can be attributed to large-scale pre-training, which helps the PLM generalize text quicker than a KN3 under low resource settings.  However, the KN3 curve has a steeper decline, which suggests that the it benefits more from the availability of additional curated corpora than a PLM in few-shot scenarios. 
We also see that the improvements produced with the additional 100x data become decreasingly pronounced with increasing train data size. i.e., once the training data is large enough to generalize the domain well, the baseline KN3 becomes a tough competitor. For HUB4, the sub-sampled KN3 even outperforms the generated  corpus for train sets larger than 20K samples. Improvements can be made by generating exponentially many samples even at ``high'' resource settings, but the cost-benefit ratio is significantly more encouraging in few-shot settings. This is in contrast to \cite{tarjan2020effectiveness}, where larger training data resulted in larger gains for word-based RNNLM models. We conclude that in domain-specific settings, although PLMs demonstrate impressive generational capabilities under few-shot conditions, n-gram models are tough baselines in comparatively higher-resourced scenarios.\vspace{-0.2cm}
\section{Conclusion}
\vspace{-0.1cm}
In this work, we compare two approaches that approximate GPT-2 into an n-gram model and find that large-scale sampling after fine-tuning performs significantly better than a probability-based approximation. We then introduce a vocabulary-restricted decoding schema that proves helpful in low-resource domain-specific scenarios. We show that interpolating the baseline trigram LM from a domain-specific corpus with trigrams made from large-scale generation improves text perplexity by 20\% across eight domains.
Using dev set perplexity to compare domains, we show that the sampling volume required to match the baseline trigram perplexity is inversely proportional to the domain modeling difficulty. We also extend text-based augmentation to few-shot scenarios, where we see impressive performance with GPT-2. However, we find that the cost benefit ratio declines rapidly with additional training data and that the trigram model is a tough baseline at higher data sizes.
\vspace{-0.1cm}
\section{Acknowledgements}
We are grateful for the feedback from Vagrant Gautam and the anonymous reviewers. Aravind Krishnan was funded by the ViCKI project within the framework of the German Civil Aviation Research Programme (LuFo VI-1) of the Federal Ministry for Economic Affairs and Climate Action (BMWK) under grant number FKZ 20D1910D.
Jesujoba Alabi was funded by the BMBF project SLIK under the Federal Ministry of Education and Research grant 01IS22015C.


\bibliographystyle{IEEEtran}
\bibliography{mybib}

\end{document}